\documentclass[acmtog,screen=true, review=false,nonacm=true]{acmart}

\usepackage{subfiles}
\def\BibTeX{{\rm B\kern-.05em{\sc i\kern-.025em b}\kern-.08emT\kern-.1667em\lower.7ex\hbox{E}\kern-.125emX}}

\setcitestyle{square}
\setcopyright{none}

\usepackage{multirow}
\usepackage{wrapfig}
\usepackage{nicefrac}
\usepackage{mathtools}
\usepackage{graphicx}
\usepackage{booktabs} 
\usepackage{combelow} 
\usepackage[ruled,vlined,linesnumbered]{algorithm2e}
\usepackage{tabularx} 
\usepackage{colortbl} 
\usepackage{manfnt} 
\usepackage{epstopdf}
\usepackage{amsmath}
\usepackage{amsfonts}
\usepackage{amsbsy}
\usepackage{verbatim}
\usepackage[mathletters]{ucs}

\definecolor{derekBlue}{RGB}{144,210,236}
\definecolor{derekTableBlue}{RGB}{189,235,252}
\definecolor{iglGreen}{RGB}{153,203,67}
\definecolor{coralRed}{RGB}{250,114,104}
\definecolor{gray}{RGB}{180,180,180}
\definecolor{orange}{RGB}{0,0,0}

\SetCommentSty{commentFont}
\SetNlSty{textbf}{}{.}

\DeclareMathOperator*{\argmax}{arg\,max}

\newcommand{\vecFont}[1]{\mathsf{#1}}

\def\vx{{\vecFont{x}}}

\def\l1{{\ell^1}}

\begin{document}

\title{Identification of stormwater control strategies and their associated uncertainties using Bayesian Optimization}

\author{Abhiram Mullapudi}
\affiliation{\institution{Xylem}
	\city{Ann Arbor}
 	\state{MI}
 	\postcode{48105}
 	\country{USA}}
\email{abhiram.mullapudi@xylem.com}
\author{Branko Kerkez}
\affiliation{\institution{University of Michigan}
 	\city{Ann Arbor}
 	\state{MI}
 	\postcode{48105}
 	\country{USA}}
\email{bkerkez@umich.edu}

\begin{abstract}
Dynamic control is emerging as an effective methodology for operating stormwater systems under stress from rapidly evolving weather patterns. Informed by rainfall predictions and real-time sensor measurements, control assets in the stormwater network can be dynamically configured to tune the behavior of the stormwater network to reduce the risk of urban flooding, equalize flows to the water reclamation facilities, and protect the receiving water bodies. However, developing such control strategies requires significant human and computational resources, and a methodology does not yet exist for quantifying the risks associated with implementing these control strategies. To address these challenges, in this paper, we introduce a Bayesian Optimization-based approach for identifying stormwater control strategies and estimating the associated uncertainties. We evaluate the efficacy of this approach in identifying viable control strategies in a simulated environment on real-world inspired combined and separated stormwater networks.
We demonstrate the computational efficiency of the proposed approach by comparing it against a Genetic algorithm. Furthermore, we extend the Bayesian Optimization-based approach to quantify the uncertainty associated with the identified control strategies and evaluate it on a synthetic stormwater network. To our knowledge, this is the first-ever stormwater control methodology that quantifies uncertainty associated with the identified control actions. This Bayesian optimization-based stormwater control methodology is an off-the-shelf control approach that can be applied to control any stormwater network as long we have access to the rainfall predictions and there exists a model for simulating the behavior of the stormwater network.
\end{abstract}

\maketitle

\section{Introduction}
%
%
The next generation of stormwater systems --- equipped with sensors and actuators --- will have the ability to autonomously control themselves to prevent flooding and improve water quality~\cite{kerkez2016}.
Over the past few years, a substantial effort has been focused on discovering algorithms for enabling the autonomous control of stormwater systems~\cite{Ocampo-Martinez_2015, lund2018,shishegar2018optimization}. 
While promising, many of these algorithms are highly parametrized, and their performance relies on finely-tuned objectives on specific study areas.
As such, the generalizability of the control methods and their applicability under uncertain conditions\footnote{e.g., rainfall uncertainty, model uncertainty, etc.} poses an open area of research.
Furthermore, the adoption of many existing control methods requires significant human and computational resources, as well as expertise in control theory, machine learning, and urban hydrology.

\

To that end, this paper, using Bayesian Optimization, proposes a generalizable and off-the-shelf approach for controlling stormwater systems and quantifying the impacts of rainfall uncertainty on control decisions.
It is ``off-the-shelf'' in that it can be used to control a stormwater system as long as there is a numerical model (e.g., SWMM, Mike-Urban) for simulating the hydraulic conditions in the stormwater network.
To the best of our knowledge, the methodology presented in this paper is the first-ever use of Bayesian Optimization to control stormwater systems.
The main contributions of this paper include:
\begin{itemize}
	\item A generalizable approach for identifying control strategies that realize watershed scale objectives.
	\item A methodology for quantifying the uncertainty associated with precipitation forecasts in the context of control decisions.
	\item An open-source implementation that can be directly adopted for the control of stormwater systems.
\end{itemize}

\section{Control of Stormwater Systems}\label{sec:ctrlstrmsystems}

Stormwater systems are designed to act as conveyance networks for transporting stormwater runoff from the urban environments into downstream waterbodies~\cite{national2009urban, rossman2010storm}.
These networks are an amalgam of interconnected storage assets, such as basins, which buffer stormwater runoff to reduce flooding and improve water quality.
While the dynamics of storm events are inherently stochastic and highly variable, stormwater networks function as static systems~\cite{kerkez2016}.
Storage and conveyance are sized for an ``average'' storm scenario, or simply governed by the availability of construction area or cost.
Once constructed, these systems are not upgraded for decades, leading to flooding and water quality impairments due to changing storm patterns and land uses. 
Simply put, many stormwater systems are unable to keep pace with rapid urbanization and evolving weather patterns~\cite{kerkez2016}.

\

Rather than rebuilding stormwater infrastructure to meet the rising demands, retrofitting existing systems with sensors and actuators poses one effective and economical alternative\cite{kerkez2016}.
Over the past decade, several simulated\cite{Mullapudi_Lewis_Gruden_Kerkez_2020, Troutman_2020, lund2018, Wong_Kerkez_2018,Ocampo-Martinez_2015,vezzaro2014} and real-world case studies\cite{Mullapudi_Bartos_Wong_Kerkez_2018} have demonstrated the effectiveness of control in mitigating flooding, reducing erosion, and improving water quality.
Control in stormwater systems enables us to tailor the network's behavior to individual storm events so that we can fully utilize the existing storage and coordinate releases to achieve watershed-scale objectives~\cite{kerkez2016}.

\subsection{State of Autonomous Stormwater Control}
The past decade has seen a significant rise in research related to autonomous stormwater control algorithms~\cite{shishegar2018optimization}.
These studies have proposed and analyzed the applicability of a wide variety of control methodologies --- ranging from static rule-based approaches to Deep learning methods like reinforcement learning.
Static rule-based\cite{schmitt2020simulation} and reactive control approaches~\cite{Troutman_2020} can be used without a surrogate model or a pre-trained controller, but designing the rules and appropriate parametrizations in these approaches requires an intimate knowledge of the stormwater network being controlled.
Wong et al.~\cite{Wong_Kerkez_2018} and others~\cite{Ocampo-Martinez_2015,joseph2014hybrid, Sun_Lorenz_2020, lund2020cso} have developed control approaches based on classical control (LQR) and linear optimization methodologies.
These approaches rely on a surrogate linear model that approximates the dynamics of stormwater network.
On the other end of the spectrum, there are Deep learning-based control approaches~\cite{Mullapudi_Lewis_Gruden_Kerkez_2020,Ochoa_Riano-Briceno_Quijano_Ocampo-Martinez_2019} which do not require a surrogate model.
But training and tuning a controller using these methods is computationally expensive~\cite{Mullapudi_Lewis_Gruden_Kerkez_2020}.
Though these approaches are very promising, adapting them for the control of new stormwater systems is a non-trivial task.
This explains, perhaps, why Genetic Algorithms-based control approaches have risen in popularity, due to their off-the-shelf nature.

\

A significant amount of research in stormwater control has relied on Genetic Algorithms (GA)~\cite{sadler2019, lund2018, Rjeily_2018, Meneses_2018, vezzaro2014}.
GA iteratively search through the set of possible solutions until converging onto a viable solution.
The biggest strength of GA is its applicability for optimizing any objective function, irrespective of its underlying mathematical structure, as long it can be numerically evaluated. 
In the context of stormwater control, GA exhaustively simulate the responses of various control decisions in a stormwater network --- using a simulator (e.g.,\ EPA-SWMM) --- until identifying a viable solution.
Though GA are intuitive and flexible for identifying control strategies that achieve a wide variety of stormwater control objectives (as long as an objective function can accurately represent them), their fundamental properties hinder their application: (i) Quality of solutions identified by GA --- especially for problems with large solution spaces like stormwater control --- is highly sensitive to the choice of hyper-parameters (e.g., population size, generations);
(ii) Irrespective of the problem, many GA randomly permutate through the solution space to identify a solution, thus not leveraging the system's inherent structure to explore the solutions efficiently; and (iii) GA are entirely opaque, and offer us no path to interpret the identified solution nor assess its optimality\footnote{Optimality in GA is evaluated through exhaustive numerical simulations}.
Readers are directed to Maier et.al for an in-depth analysis on the use GA in the water systems~\cite{maier2014}.

\

Furthermore, recent works in the control of stormwater systems have also acknowledged the limitation of the existing approaches in quantifying the rainfall uncertainty associated with control decisions\cite{sadler2019}.
Hence, there is a need for an easy to use and generalizable control approach, that addresses the limitations of GA and can quantify uncertainty.
Such an approach would be instrumental in transitioning autonomous stormwater control algorithms from simulation into adoption on physical systems.
The Bayesian Optimization-based control approach presented in this work is one such approach.

\section{Control of stormwater systems using Bayesian Optimization}\label{sec:bae}
\begin{figure}
	\centering
	\includegraphics[width=0.90\linewidth]{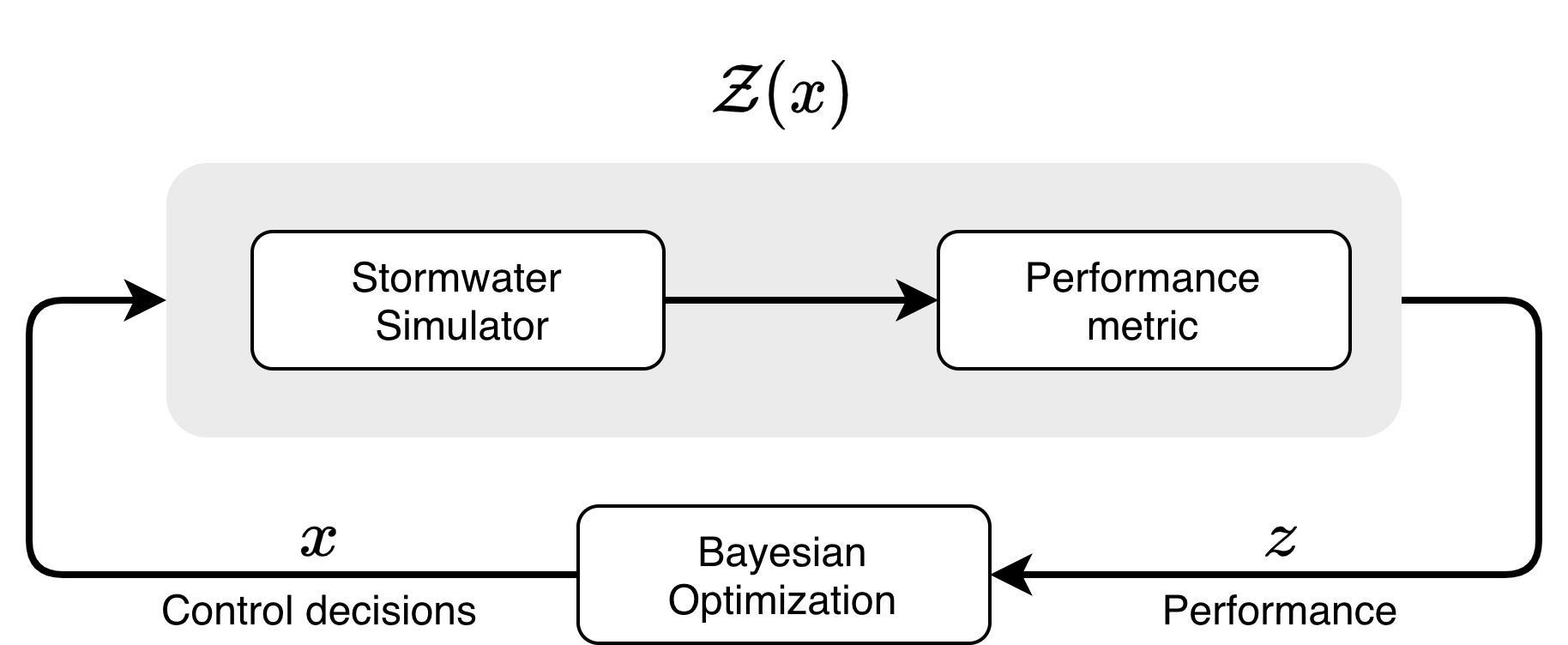}
	\caption{Bayesian Optimization converges onto an optimal control decision that realizes the stormwater network's desired response by learning a surrogate objective function and identifying the control decision that minimizes it. Surrogate objective functions, in Bayesian Optimization, maps the control decisions ($\vx$) to their corresponding performance ($z$). It is learned by evaluating the performance of the various control decisions in the solution space. A control decision's performance is evaluated by simulating its response using a stormwater simulator and quantifying the simulated response (e.g.\ hydrographs, water levels, and pollutographs) using a performance metric.}\label{fig:bae}
\end{figure}

Bayesian Optimization (BO) is used optimize systems that are computationally expensive to simulate and whose dynamics restrict them from being easily formulated in the classical optimization framework~\cite{frazier2018tutorial}.
BO optimizes systems by learning a surrogate objective function.
It relies on this function to quantify the uncertainty associated with the identified optimum and reduce the computational resource required for the optimization process~\cite{frazier2018tutorial}. 
BO is extensively used in the Artificial Intelligence community to optimize hyper-parameters in Deep Learning models~\cite{chen_huang_2018, brochu2010tutorial, frazier2018tutorial}.
In the water domain, Candelieri et al.\ used BO for pump scheduling optimization in water distribution networks~\cite{Candelieri_Perego_Archetti_2018}.
To the best of our knowledge, BO has not yet been applied for the control of stormwater systems.

\

Stormwater control strategies have relied on reactive and horizon-based control approaches, which control the network's controllable assets throughout the storm event~\cite{sadler2019, vezzaro2014, Troutman_2020, Wong_Kerkez_2018}.
This paper uses a control strategy that requires configuring the controllable assets in the stormwater network only \textit{once} before the storm event to achieve the desired flow control objectives.
Such a control strategy reduces the number of control interventions\footnote{Number of times a valve opening or a pump setting has to be modified during a storm event} that are required for controlling a system. 
Unlike feedback control, in this proposed planning based control approach, the controller does not alter its course if its actions result in an unintended response.
Using BO, based on the existing numerical model (e.g.\ EPA-SWMM) and anticipated storm event, we identify the control decisions that should be implemented in the stormwater network to realize the desired response.
Hence, planning for the possible uncertainties and choosing a control decision that minimizes the risk becomes essential.
BO's ability to quantify uncertainties associated with control decisions makes it ideal for such a control strategy.

\

Stormwater dynamics are governed by non-linear phenomena, such as runoff, infiltration, and gravity-driven flow through complex networks of basins and channels~\cite{rossman2010storm, Mullapudi_Wong_Kerkez_2017, Rimer2019}.
Prior studies~\cite{lund2020cso} that formulated the control of stormwater systems as an optimization problem were constrained to approximating these phenomena as a linear system.
This is a non-trivial task, not to mention one that limits the analysis of important higher-order non-linear terms. 
Our BO formulation can optimize systems by merely relying on a numerical model without requiring a surrogate model or simplified dynamics.
BO can be adopted as long as there is a way (e.g.\ numerical simulation, empirical methodology, or a real-world study) to evaluate the impacts of a control decision~\cite{frazier2018tutorial}.
Many popular models for stormwater exist and are used by communities and municipalities worldwide (e.g.\ SWMM, MIKE)~\cite{rossman2010storm}.
Our approach also builds upon what many city managers already use --- thus providing, as much as possible, ``off-the-shelf'' approach for control of stormwater systems.

\

We formulate the control of stormwater systems as an optimization problem, in which the objective function ($\mathcal{Z}$), subject to constraints, captures the desired response of a stormwater system to control actions. 
The objective function in BO is evaluated by simulating the control decision's response in a stormwater simulator and then distilling the simulated hydrographs and water levels into a \textit{performance metric} that quantifies the degree of success of the control decision.
Consider the scenario where we want to identify the valve position at an outlet structure to maintain the outflows from the basin below a threshold.
This control objective's performance metric is formulated in Eq.~\ref{eq:opti-ctrl}, as the cumulative flows exceeding the threshold ($\lambda$) during the storm event for an outlet valve position, constrained between completely closed (0.0) to completely open (1.0).
\begin{subequations}\label{eq:opti-ctrl}
\begin{alignat}{2}
&\!\arg\min_{x}        &\qquad& \mathcal{Z}(\vx) = \sum_{t=0}^{T} g(q_t) \\
&\text{subject to} &      & 0.0 \leq \vx \leq 1.0
\end{alignat}
\begin{equation}
	g(q_t) = \begin{cases}	
		(q_t - \lambda) &\text{if } q_t > \lambda\\
		0.0 &\text{else }\\
	\end{cases}
\end{equation}
\end{subequations}
$T$ and $q_t$ in Eq.~\ref{eq:opti-ctrl} represent the storm event duration and the outflows from the basin.
The valve position that achieves the desired flow control objective is determined by identifying the control decision $(\vx)$ that minimizes the performance metric in Eq.~\ref{eq:opti-ctrl}.

\

BO  learns the surrogate objective function that maps the control decisions to their corresponding performance metrics ($\mathcal{F}:\vx \rightarrow z$) and then uses it to identify the optimal control decision~\cite{frazier2018tutorial}.
This surrogate objective function is learned by evaluating the given objective function's value across various possible control decisions.
Gaussian Processes (GP) are the most widely used regressors in BO for learning this mapping as they, for a given set of data-points, learn to predict the performance metric associated with a control decision and estimate the uncertainty associated with the prediction~\cite{rasmussenGaussianProcessesMachine2006, frazier2018tutorial}.
The most distinguishing characteristic of BO is its ability to prioritize the most promising solutions when searching through the solution space, using an acquisition function ($\mathcal{A}$), to minimize the number of data-points --- and subsequently the number stormwater simulations --- required for identifying the optimum~\cite{rasmussenGaussianProcessesMachine2006}.

\begin{algorithm}
\caption{Bayesian Optimization approach for controlling stormwater systems. Let $n_o$ be the number of initial random eval\-uations in the solution space and $N$ be the total number of evaluations.}\label{algo:bayes}
Evaluate the performance of $n_o$ randomly sampled control decisions;  $\mathcal{D}_{0:n_o}:\{\vx_{0:n_o}, z_{0:n_o}\}$\\
Learn an initial estimate of the surrogate objective function using a GP; $\mathcal{F}_{n_o} \sim \mathcal{GP}(\mathcal{D}_{0:n_o})$\\
Set n = $n_o$\\
\While{$n \leq N$}{Identify a new control decision to evaluate; $\vx_{n+1} = \arg\max \mathcal{A}(x |\mathcal{F}_n)$ \\
	Evaluate performance ($z_{n+1}$)  of $\vx_{n+1}$ using a stormwater simulator.\\
	Augment $\mathcal{D}_n$ with $\{ \vx_{n+1}, z_{n+1}\}$\\
	Update $\mathcal{F}_{n+1} \sim \mathcal{GP}(\mathcal{D}_{0:n+1})$\\
	n = n+1
}
Return $\arg\min \mathcal{F}_N(x)$
\end{algorithm}

\

Algorithm.~\ref{algo:bayes} summarizes the BO-based control of stormwater systems.
Initially, BO evaluates a pre-defined number ($n_o$) of random control decisions, in the solution space, to create an initial set of data points ($\mathcal{D}_{0:n_o}:\{ \vx_{0:n_o}, z_{0:n_o} \}$). 
These are used to learn an initial estimate of the surrogate objective function ($\mathcal{F}$), which then is used by the acquisition function ($\mathcal{A}$) for identifying the next control decision ($\vx_{n+1}$) to evaluate.
Using the performance metric, we evaluate this control decision's ability to achieve the desired response and update the set of data points with these observations ($\{ \vx_{n+1}, z_{n+1}\}$).
These new set of data-points are then used to update the surrogate objective function.
This updated function will be used by the acquisition function in the next iteration to identify a new control decision to evaluate.
This process is repeated until convergence onto an optimum or for a pre-defined number of iterations.

\

\begin{figure*}
	\includegraphics[width=\textwidth]{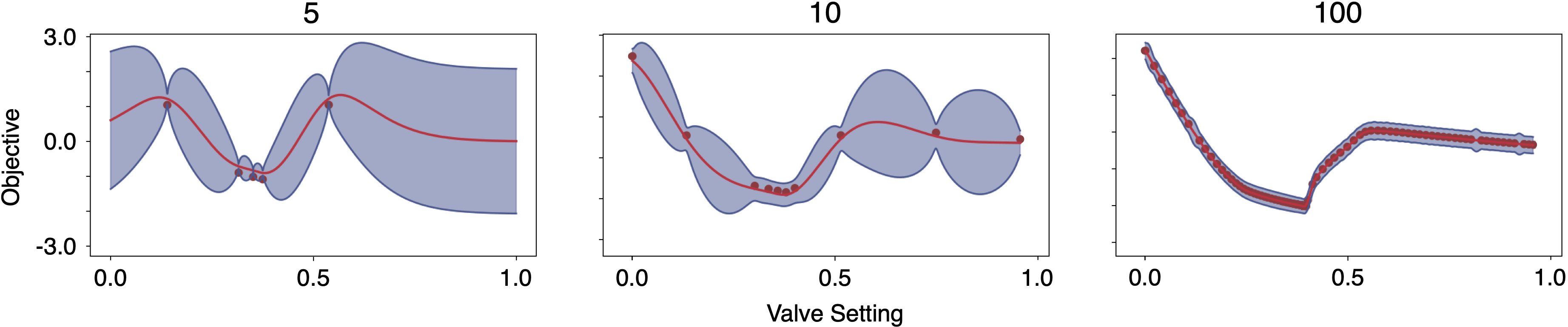}
	\caption{With every iteration of BO, the uncertainty (shaded area) in the surrogate objective function reduces. This figure illustrates the surrogate objective function being approximated by the GP at various iterations. Initially (N=5), there is a high degree of uncertainty in the GP's estimate of control decision's performance. As the number iterations increase, the uncertainty reduces. Also, notice that the BO focuses its initial evaluations  (N=5 and 10) in the most promising regions of the solution space.}\label{fig:single-bayopt}
\end{figure*}

Every iteration in BO improves the estimate of the surrogate objective function. 
Fig.\ref{fig:single-bayopt} illustrates the surrogate objective function across various iterations (5, 10, 100) learned by the BO approach identifying the control action that maintains the outflows from the basin below a threshold (Eq.~\ref{eq:opti-ctrl}).
In Fig.\ref{fig:single-bayopt}, red dots are the evaluated control decisions, and the red line and shaded area represent the GP estimate of the surrogate objective function and its associated uncertainty. 
As the number of the iterations increase, the uncertainty associated with the surrogate objective predictions decreases.
Based on the initial evaluations, the acquisition function identifies that there is a high probability that the optimum might lie between 0.3 and 0.5 and focuses its search in this region (indicated by the high density of red dots in Fig.\ref{fig:single-bayopt}).

\

In the following sections, we present an overview of GP and acquisition functions and introduce an algorithm for updating the GP in BO to provide more accurate uncertainty estimates.

\subsection{Gaussian Processes}\label{subsec:gp}
BO relies on GP to learn the surrogate objective function that maps the control decision ($\vx$) to the performance metric  ($z$) that represents the control decision's ability to achieve the desired objective successfully~\cite{frazier2018tutorial}. 
GP assumes the standard regression model (Eq.~\ref{eq:gp-regres}), where the estimate of the performance metric is a function of the control decision and uniform Gaussian noise ($\epsilon \sim \mathcal{N}(\mu_n, \sigma_n^2)$)~\citep{rasmussenGaussianProcessesMachine2006}.
\begin{equation}\label{eq:gp-regres}
	z = f(\vx) + \epsilon
\end{equation}
Apart from the Gaussian assumption on the noise, GP does not require any further assumptions on the structure of the mapping or the data's nature.
When trained, GP not only learn to predict a value but also estimate the uncertainty associated with the prediction~\cite{rasmussenGaussianProcessesMachine2006}.  
GP's ability to quantify uncertainties and their flexibility to be applied to any data without requiring any explicit assumptions makes GP one of the most efficient and generalizable approach for regression, especially for hydrological systems. 
GP have been used by Fries et al.\ to improve the accuracy of hydrological models~\cite{Fries_Kerkez_2017}, and Troutman et al.\ have used GP for identifying the contribution of wastewater flows in combined sewer systems~\cite{Troutman_Schambach_Love_Kerkez_2017}.

\

A GP is defined as a collection of random variables, any finite subset of which are jointly Gaussian~\cite{rasmussenGaussianProcessesMachine2006}.
In this context, the control decision's performance ($f(\vx)$) is the random variable, and they are assumed to be jointly Gaussian (i.e.\ $P(f(\vx_{n+1})|f(\vx_1)f(\vx_2) \ldots f(\vx_{N})) \sim \mathcal{N}(\mu,K))$.
For example, let say we evaluated the performance of $N$ control decisions, and we wish to know the performance of an unevaluated control decision ($\vx_{N+1}$).
GP relying on the jointly Gaussian property, predict the performance value at the unevaluated control decision $(z_{N+1})$ by identifying the mean $(\mu_{1:N})$ and covariance $(K_{1:N})$ that characterizes the underlying distribution in the observed data~\cite{rasmussenGaussianProcessesMachine2006}.

\

A GP is fully characterized by a mean function and a covariance function (Eq.~\ref{eq:GP}).
The mean function, given a set of observed data points ($\mathcal{D}_{N}:\{ \vx_{0:N}, z_{0:N}\}$), predicts the most likely value ($z$) at the data point  and covariance function estimates the uncertainty associated with this prediction.
\begin{align}\label{eq:GP}
	f(\vx|\mathcal{D}_N) &\sim \mathcal{GP} ( \mu(\vx|\mathcal{D}_N) , K( \vx|\mathcal{D}_N) ) 
\end{align}
Fig.~\ref{fig:single-bayopt}  summarizes the mean function's predictions (indicated by the red line) and the covariance function's uncertainty estimates (95\% confidence interval indicated by the shaded area) in the solution space $([0,1])$, as the set of observed data points (indicated by the red dots) increases.
In Fig.~\ref{fig:single-bayopt}, the mean function's predictions at observed data points $(\vx_{0:N})$ align perfectly with the observed values (i.e.\ $\mu(\vx_{0:N}) = z_{0:N}$), and the estimated uncertainty at these points is almost 0.
Uncertainty estimates for the unobserved data points in Fig.~\ref{fig:single-bayopt} increase as they get further from the observed data, indicating that the values predicted by the mean function at these points are inaccurate.

\

The predictions of GP for a set of data points are determined by their similarity (or dissimilarity) to the collection of observed data points.
This similarity in GP --- specifically in the mean and covariance functions --- is quantified using kernels.
The governing physical relationships between control decisions and their impacts in stormwater systems\footnote{e.g.\ Outflow from a valve in a basin is $C_d \times \text{valve opening} \times \sqrt{2 \times g \times h}$} are non-linear, and kernels enable us to embed these non-linear relationships.
Squared Exponential Kernel (Eq.~\ref{eq:kenel-sqexp}) is widely used in the GP literature for capturing such non-linear relationships~\cite{Duvenaud}.
\begin{equation}\label{eq:kenel-sqexp}
	k(\vx, \vx^\prime) = \sigma^2 \exp 	\big( {-\frac{{(\vx - \vx^\prime)}^2}{2{l}^2}} \big)
\end{equation}
Please refer to Rasmussen et al.\ for more information on kernels~\cite{rasmussenGaussianProcessesMachine2006}. 
Consider the control of an outlet valve in a basin, where evaluated the performance ($z$) of the valve opening at 50\% ($\vx=0.50$), and we wish to know the performance ($z^\prime$) at 60\% opening ($\vx^\prime=0.60$). 
Using GP, it can be analytically estimated using the following equations:
\begin{subequations}\label{eq:me-cov}
\begin{align}	
	\mu(\vx^\prime) &= k(\vx^\prime, \vx){[k(\vx, \vx) + \sigma_n^2]}^{-1}z \\
	K(\vx^\prime) &= k(\vx^\prime, \vx^\prime) - k(\vx^\prime, \vx){[k(\vx, \vx) + \sigma_n^2]}^{-1}k(\vx, \vx^\prime)\\
	z^\prime &\sim \mathcal{N}(\mu(\vx^\prime), K(\vx^\prime))
\end{align}
\end{subequations}

\

A final decision of the GP chain involves finding the choice of hyper-parameters ($ \theta : \{ \sigma^2, l, \sigma_n^2 \}$) that best represent the observed data points.
In this paper, these parameters are determined by maximizing the log marginal likelihood~\cite{rasmussenGaussianProcessesMachine2006}.
\begin{subequations}\label{eq:log-like}
\begin{align}
	\log p(\mathbf{z} | X, \theta) &= -\frac{1}{2} \mathbf{z}^{\top} K_{z}^{-1} \mathbf{z}-\frac{1}{2} \log \left|K_{z}\right|-\frac{n}{2} \log 2 \pi \\
	K_z &= K_{N \times N} + \sigma_n^2 I_{N \times N}
\end{align}
\end{subequations}
Log marginal likelihood (Eq.\ref{eq:log-like}) is a metric that quantifies the accuracy with which a given set of model parameters represent the data.
$K_{N \times N}$ and $I_{N \times N}$ in Eq.~\ref{eq:log-like} represent the kernel matrix and the identity matrix.

\subsection{Acquisition Function}
The acquisition function, based on the surrogate objective function learned by the GP, identifies the most promising control decisions to evaluate in the solution space.
BO relies on the acquisition function to efficiently explore the solution space and minimize the number of evaluations required for converging onto an optimum~\cite{frazier2018tutorial}.
Expected Improvement (EI), one such widely used acquisition function in BO~\cite{jones1998efficient, frazier2018tutorial}, is defined as follows:
\begin{equation}\label{eq:EI}
		EI_n(x) = E \big[ {[ f_n(x) - f^*_n ]}^+ \big]
\end{equation}
Where $x$ and $f_N(x)$ are the control decision and its corresponding value estimated using the surrogate objective function learned from $N$ data points.
$f^*_N = \max f(x)$ is the current best performing solution. 
${[ f_N(x) - f^*_N ]}^+$ indicates the positive part (i.e.\ $z^+ = \max (z, 0)$).
\begin{equation}\label{eq:ei-max}
	x_{N+1} = \argmax EI_N(x)
\end{equation}
By identifying the control decision that maximizes the EI, BO identifies the next control decision $x_{N+1}$ to evaluate~\cite{frazier2018tutorial}. 

\

The analytical nature of the GP enables us to derive a closed-form solution (Eq.\ref{eq:cloase-ei}) for EI~\cite{jones1998efficient}.
This closed-form equation is then used as the objective function in Eq.\ref{eq:ei-max} to identify the next control decision to evaluate.
\begin{align}\label{eq:cloase-ei}
	EI_N(x) &= {[\Delta_N(x)]}^+ + \sigma_N(x) \varphi(\frac{\Delta_N(x)}{\sigma_N(x)}) - |\Delta_N(x)| \Phi(\frac{\Delta_N(x)}{\sigma_N(x)})
\end{align}
$\mu_N(x)$ and $\sigma_N(x)$ in Eq.\ref{eq:cloase-ei} are the mean and covariance function of the GP.
$\varphi$ and $\Phi$ are the PDF and CDF of standard normal distribution.
$\Delta_N(x) = \mu_N(x) - f^*_N$ is the difference between the performance of a suggested control decision $x$ and the best performing solution in the surrogate objective function.
Previous studies~\cite{frazier2018tutorial} have indicated that the use of numerical optimization methods like quasi-Newton method have worked well in optimizing Eq.\ref{eq:cloase-ei}.
Please refer to \textit{A Tutorial on Bayesian Optimization} by Peter I. Frazier for a detailed discussion on EI and other ``exotic'' acquisition functions that are being used in BO~\cite{frazier2018tutorial}.

\subsection{Uncertainty Quantification}
GP described in Section-\ref{subsec:gp}, when learning the surrogate objective function, assume a uniform noise ($\sigma^2_n$) on control decision's performance.
Under this assumption, GP optimizing for the best set of parameters identify a $\sigma^2_n$ that encapsulates the entire set of observed data~\cite{rasmussenGaussianProcessesMachine2006, Kersting_Plagemann_Pfaff_Burgard_2007}.
Consider the instance where a controlled stormwater basin can experience a storm event in an ensemble of probable events.
Under these uncertain conditions, using BO, we want to identify the best control decision that we can implement in the basin and quantify the uncertainty that this control decision would achieve the desired objective.
GP trained with the assumption of uniform noise tends to over or underestimate the risks associated with individual decisions when the variance in the performance depends on the decision~\cite{Kersting_Plagemann_Pfaff_Burgard_2007}, which is the case for the stormwater systems.
For example, if the outlet valve in the stormwater basin is opened by 0.10\% and the basin experiences two different storm events --- one small and one large --- the variance in the outflows leaving the basin is low, rather than when the valve is opened by 0.90\%.

\
\begin{algorithm}
\caption{Uncertainty Quantification using MLH-GP}\label{algo:hetrosodastic-GP}
Let $\mathcal{GP}_1$ be the GP representing the surrogate objective function ($\mathcal{F}$) learned by the Bayesian Optimizer in Algorithm.~\ref{algo:bayes}\\
Given $\mathcal{GP}_1$, estimate the empirical noise ($\hat{z}$) from the training data $\hat{z}_i = \log(var[z_i, \mathcal{GP}_1(\vx_i|\mathcal{D})])$ and create a new data set $\hat{\mathcal{D}_{0:N}} = \{ x_{0:N}, \hat{z}_{0:N} \}$ \\
Using $\hat{\mathcal{D}}$, learn $\mathcal{GP}_2$ to estimate noise ($\hat{z}$) from $\vx$; $\mathcal{GP}_2: \vx \rightarrow \hat{z}$ \\
Using $\mathcal{GP}_2$ to estimate logarithmic noise, train a combined $\mathcal{GP}_3: \vx \rightarrow z$ using Eq.~\ref{eq:hgp-gp3} and Eq.~\ref{eq:gp3-cov}\\
If not converged, set $\mathcal{GP}_2$=$\mathcal{GP}_3$ and go to step 2 \\
\end{algorithm}

Unlike the generic GP regression model, which assumes uniform Gaussian noise (Eq.\ref{eq:gp-regres}), Most Likely Heteroscedastic Gaussian Process (MLH-GP) builds on the assumption that the Gaussian noise in the regression model is dependent on its inputs~\cite{Kersting_Plagemann_Pfaff_Burgard_2007}.
\begin{subequations}\label{eq:mlhgp-regress}
\begin{align}
	z_i &= f(\vx_i) + \epsilon_i \\
	\epsilon_i &\sim \mathcal{N}(0, r(\vx_i)) 
\end{align}
\end{subequations}
This regression model (Eq.\ref{eq:mlhgp-regress}) enables us to account for the variance in the performance stemming from each control decision.
We can directly use MLH-GP to update the GP ($\mathcal{GP}_1$) used for learning the surrogate objective function in BO to account for these variances. 
\begin{subequations}\label{eq:hgp-var}
\begin{align}
	\hat{z}_i &= var[z_i, \mathcal{GP}_1(\vx_i|\mathcal{D})] = {s}^{-1} \times \sum^s_{j=1} 0.5 \times {(z_i - z_i^j)}^2 \\
	z_i^j &\sim \mathcal{N}(\mu_{\mathcal{GP}_1}(\vx_i), K_{\mathcal{GP}_1}(\vx_i))
\end{align}
\end{subequations}
MLH-GP, using Eq.\ref{eq:hgp-var}, estimate the variance in the performance ($z_i$) observed by simulating the control decision ($\vx_i$) and performance predicted ($z_i^j$) by the learned surrogate objective function ($\mathcal{GP}_1$).
A new GP ($\mathcal{GP}_2$) is then trained to learn the mapping between the simulated control decisions and log of estimated variance.
$s$ in Eq.\ref{eq:hgp-var} is the number of samples used for estimating variance~\cite{Kersting_Plagemann_Pfaff_Burgard_2007}.
\begin{subequations}\label{eq:hgp-gp3}
	\begin{align}
		r(x) &= e^{\mu_{\mathcal{GP}_2}(x)}\label{eq:nos-esti}\\
		R(x) &= diag(r(x))
	\end{align}
\end{subequations}
Using the estimates of $\mathcal{GP}_2$, we populate the $R$ matrix (Eq.~\ref{eq:hgp-gp3}) and then use this matrix to train a new GP ($\mathcal{GP}_3$) that accounts for the input dependent variances (Eq.~\ref{eq:gp3-cov}).
In this paper, we use this GP to quantify the uncertainty associated with control decisions.
\begin{subequations}\label{eq:gp3-cov}
\begin{align}
	\mu_{\mathcal{GP}_3} &= K^{*}{(K+R)}^{-1} \mathbf{z}\\
	K_{\mathcal{GP}_3} &= K^{* *}+R^{*}-K^{*}{(K+R)}^{-1} K^{* T}\\
	\mathcal{GP}_3 &\sim \mathcal{N} (\mu_{\mathcal{GP}_3}, K_{\mathcal{GP}_3})
\end{align}
\end{subequations}
As described in the Algorithm.~\ref{algo:hetrosodastic-GP}, we repeat this process until the estimates of $\mathcal{GP}_3$ converge.

\section{Evaluating Bayesian Optimization for the control of stormwater systems}\label{sec:methods}
Stormwater networks are complex and often designed uniquely to meet the requirements of a particular watershed.
During their operation, these systems experience a wide spectrum of storm events, many of which they might not have been explicitly designed for.
These systems are required to achieve a multitude of localized and system scale objectives.
Hence, for a stormwater control approach to be applicable, it has to achieve a diverse set of control objectives across various network topologies and storm events.
Furthermore, control algorithms should have the ability to account for the uncertainties --- especially those stemming from the stochastic nature of storm events --- that are inherent in these systems. 
Though a real-world evaluation is essential for validating a control approach's applicability, we focus on a simulation-based evaluation, as it will allow us to exhaustively simulate a number of scenarios.

\

In this paper, we analyze the BO's ability to control stormwater systems across two criteria:
\begin{enumerate}
	\item \textbf{Generalizability}: We analyze the applicability of BO for achieving diverse control objectives across stormwater systems and rain-events.
	\item \textbf{Uncertainty Quantification}: We evaluate BO's ability to quantify rainfall uncertainty associated with a particular control decision.
\end{enumerate}
We assess these criteria using the stormwater control scenarios from \texttt{pystorms}, an open-source python package developed for the quantitative evaluation of stormwater control algorithms.
\texttt{pystorms} provides a curated collection of stormwater networks, storm events, and control objectives delineated as scenarios, named after the Greek alphabet, for evaluating stormwater control algorithms~\cite{Rimer2019}.
In the following Sections-~\ref{sec:general} and~\ref{sec:unq}, we specify the specific scenarios and metrics used for quantifying the ability of BO in each of the above-described criteria.
In Section--~\ref{sec:implementation} , we summarize the computational implementation of these evaluations.

\subsection{Generalizability}\label{sec:general}
We quantify the Generalizability of BO by analyzing its ability to identify the control decisions that achieve the desired objective in both separated and combined stormwater systems.
Specifically, scenario gamma and epsilon in \texttt{pystorms}.
In this control approach, we implement only a control decision for the entire duration of the storm event. Hence, we limit our evaluation to event-specific control objectives, with the specific goals of minimizing flooding and maintaining flows and loading below a threshold.
This evaluation assumes that we have perfect information about the storm event (inputs) used to drive the stormwater network's response.

\begin{subequations}\label{eq:gen-gamma-perf}
\begin{align}
	\mathcal{Z} =  \overbrace{\sum_i^N c_1 \times d^i_T}^{\text{Storage penality}} &+ \sum^T_t \sum_i^N \Big\{ \overbrace{\frac{c_2 }{T} \times f^i_t}^{\text{Flooding penality}} + \overbrace{g(q^i_t)}^{\text{Flow penalty}} \Big\}\\
	g(q^i_t) &= \begin{cases}	
		(q^i_t - 0.11) &\text{if } q^i_t > 0.11\\
		0.0 &\text{else }\\
	\end{cases}
\end{align}
\end{subequations}

\

Scenario gamma represents a separated stormwater network in a semi-urbanized watershed, and this network comprises 11 interconnected basins that are draining into a downstream water body.
Equipped with a controllable valve, these basin's outlet size can be adjusted to any value between 100\% open to completely closed.
The control objective in this scenario is to maintain the flows in the network below $0.11 m^3\ s^{-1}$ during a 25 year 6-hour SCS-II design storm and is quantified based on the performance metric presented in Eq.~\ref{eq:gen-gamma-perf}, where $c_1=10^3, c_2=10^4$ represent the penalty factors.
These penalties are determined such that they penalize flooding more than storing water in the basins.
$d_T$, $f_t$, and $q_t$ in Eq.~\ref{eq:gen-gamma-perf} represent depth, flooding and outflows during the simulation.
The storage component penalizes the controller for storing water in the basins at the end of the storm event.
This penalty is incorporated in the performance metric to prevent the controller from converging onto a trivial solution that stores all the runoff in the basins to maintain flows below the desired threshold.
The flow and flooding penalty penalize the controller based on the cumulative flooding and flows that exceed the desired threshold of $0.11 m^3\ s^{-1}$.
Hence, by attempting to identify a control decision that minimizes this performance metric, the controller is compelled to identify the set of valve positions (i.e.\ the percent of outlet opening) that maintain the network's flows below the desired threshold, while avoiding flooding.
In this analysis, we control the four most downstream basins, chosen based on their control potential, in the network~\cite{Mullapudi_Lewis_Gruden_Kerkez_2020, schutze_sewer_2008}. 


\begin{subequations}\label{eq:gen-epslion-perf}
\begin{align}
	\mathcal{Z} =  \sum^T_t \big\{ \overbrace{g(l_t)}^{\text{Loading penalty}} &+ \sum_i^N  \overbrace{h(f_t^i)}^{\text{Flooding penality}} \big\}\\
	g(l_t) &= \begin{cases}	
		(l_i - 1.075) &\text{if } l_t > 1.075\\
		0.0 &\text{else }
	\end{cases}\\
	h(f^i_t) &= \begin{cases}	
		10^9 &\text{if } f^i_t > 0.0\\
		0.0 &\text{else }
	\end{cases}
\end{align}
\end{subequations}

\

Scenario epsilon represents a combined stormwater system in an urban watershed of 67$km^2$, comprising a network of pipes draining, both wet and dry weather flows, into a downstream water treatment plant.
This network has 11 inflatable dams that can hold back (or release) runoff in the pipes.
This effectively seeks to create interim storage for the runoff in the pipe network, before those flows reach a downstream treatment plant. 
The control objective in the scenario is to maintain the loading at the network's outlet below $1.075\ kg\ s^{-1}$ while preventing flooding. 
This scenario is driven by three (two of which are back-to-back) real-world storm events and daily diurnal flows.
The control decision's ability to realize this control objective is quantified using the performance metric presented in Eq.~\ref{eq:gen-epslion-perf}.
Loading penalty in the performance metric penalizes the control actions resulting in exceedance of loads (i.e.\ $>1.075\ kg\ s^{-1}$) leaving the network, and the flooding penalty ensures that these control actions avoid flooding.
In this analysis, we control all the 11 inflatable dams.

\

In this analysis, we also evaluate the GA's ability to identify the optimal control decisions for the above described two scenarios and compare its performance to the BO approach to analyze their ability to identify a better performing solution\footnote{(i.e.\ identify the set of control decisions that minimize the metrics defined in Eq.~\ref{eq:gen-epslion-perf} and Eq.~\ref{eq:gen-gamma-perf})} for the same number of stormwater simulations.
Specifically, we compare the optimal solutions identified by the BO and GA approach for 30 stormwater simulations.
Given the stochastic nature of these approaches, we repeat this evaluation for 30 different random seeds and compare the mean and variance of the performance of identified solutions.

\subsection{Uncertainty Quantification}\label{sec:unq}
We evaluate BO's ability to quantify the rainfall uncertainty by comparing its uncertainty estimates to empirically computed values.
For this evaluation, we consider the scenario where the stormwater system can experience any one storm event, with uniform probability, from the ensemble of possible events.
Uncertainty in this evaluation is empirically computed by evaluating  each possible action's performance (Eq.~\ref{eq:perf-unq}) across all the possible storm events and computing its mean and variance so that they can be compared to the values estimated by the GP in BO.
\begin{subequations}\label{eq:perf-unq}
\begin{align}
	\mathcal{Z} &= - \exp\{ \sum_{t=0}^T  \frac{f_t}{i_t} + \frac{g(q_{t})}{i_t} \} \\
	g(q_{t}) &=
      \begin{cases}
        q_{t} - 1.0
                   & \text{if \,} q_{t} > 1.0 \\
	      0.0     & \text{else}
      \end{cases}
\end{align}
\end{subequations}
$f_t$, $i_t$, and $q_t$ in the performance metric represent flooding, inflow, and outflows during the simulation.
Unlike the other two scenarios, the performance metric for uncertainty quantification is defined as an exponential function as it creates a smoother terrain for the MLH-GP to approximate.
Krauth et al. have argued that a well-defined objective function is essential for the GP to approximate a function accurately~\cite{Krauth_Bonilla_Cutajar_Filippone}.

\

We focus our evaluation on a stormwater system with a stormwater basin (scenario theta in \texttt{pystorms}), equipped with a controllable valve at its outlet, experiencing a set of 20 synthetically generated storm events.
This scenario's control objective is to maintain the outflows from the basin below $1.0\ m^3 s^{-1}$ for any/all of the possible storm events.
We limit our evaluation to the control of a single basin. As its solution space is small enough that an empirical evaluation of all possible solutions is computationally tractable\footnote{1\% discretization of solution space from 0 to 100\% for 20 storm event would require 2000 simulations.}.
We embed the notion of uncertainty in rainfall by randomizing the storm event in the stormwater simulator when evaluating the performance of a control decision. 
For every iteration of the BO, we uniformly sample a storm event from the possible set of storm events and simulate its response for the control decision.  
Then the surrogate objective function (i.e., GP) is updated using Algorithm.~\ref{algo:hetrosodastic-GP}. 
We then compare these updated estimates to the empirically computed values to evaluate their accuracy. 
Furthermore, to analyze the effectiveness of MLH-GP in quantifying uncertainty estimates, over GP, we evaluate the performance of control actions sampled uniformly across the solution space and train these GPs to estimate the uncertainties in these performance estimates.

\subsection{Implementation}\label{sec:implementation}
The scenarios in \texttt{pystorms} are equipped with a stormwater simulator, which can be accessed using a Python-based programming interface.
This Pythonic interface enables us to integrate the Python-based scientific computing libraries with the stormwater simulator and develop advanced control algorithms, such as the one presented in this paper. 
\texttt{pystorms} uses US EPA's Stormwater Management Model (SWMM), an extensively used open-source stormwater modeling software~\cite{rossman2010storm}, as its stormwater simulator.
SWMM is built in C language and interfaces with \texttt{pystorms} via \texttt{PySWMM}, a python wrapper that enables us to use the functionality of SWMM from Python~\cite{mcdonnell_bryant_e_2020_3751574}.
SWMM uses a dynamical wave equation for routing runoff through the stormwater network, which accounts for backchannel flows and other such dynamics that can arise in the network from control, thus making SWMM an ideal choice for simulating control of stormwater systems~\cite{Mullapudi_Wong_Kerkez_2017}. 
Please refer to Rimer et al.\ for more information on \texttt{pystorms} and an overview of modeling control in stormwater systems~\cite{Rimer2019}.

\

BO implementation has two primary components, GP and Acquisition functions. 
As alluded in Section-~\ref{subsec:gp}, there are a lot of moving parts in implementing GP~\cite{frazier2018tutorial}.
Over the years, a suite of accessible open-source libraries has been developed that, right out of the box, provide implementations of various kernels and automate the process of identifying parameters to prototype the GP with minimal overhead~\cite{gpy2014, scikit-learn, daulton2020differentiable}.
\texttt{GPy} developed by Sheffield machine learning group is one such Python library.
In this paper, we have used \texttt{GPy} for modeling GP and MLH-GP~\cite{gpy2014}. 
We use \texttt{GPyOpt}, another Python library developed Sheffield machine learning group, for implementing Bayesian Optimization~\cite{gpyopt2016}. 
\texttt{GPyOpt} provides an implementation of the Acquisition functions that build on the GP implemented using \texttt{GPy} and a modular programming interface, used in this paper to interface with \texttt{pystorms}.
In this work, GA evaluation is implemented using DEAP, one of the most widely used Python-based evolutionary computation framework\cite{fortin12a}. Specifically, we adopt the implementation of One Max Problem provided in the library~\cite{fortin12a}.
Simulations presented in this paper were run in the University of Michigan's Great Lakes Custer on a 3.0 GHz Intel Xeon Gold 6154 processor with 1GB of RAM and the source code used for this evaluations can be accessed at \href{https://github.com/kLabUM/BaeOpt}{github.com/kLabUM/BaeOpt}.

\section{Results}
\subsection{Generalizability}

\begin{figure*}
	\includegraphics[width=0.80\linewidth]{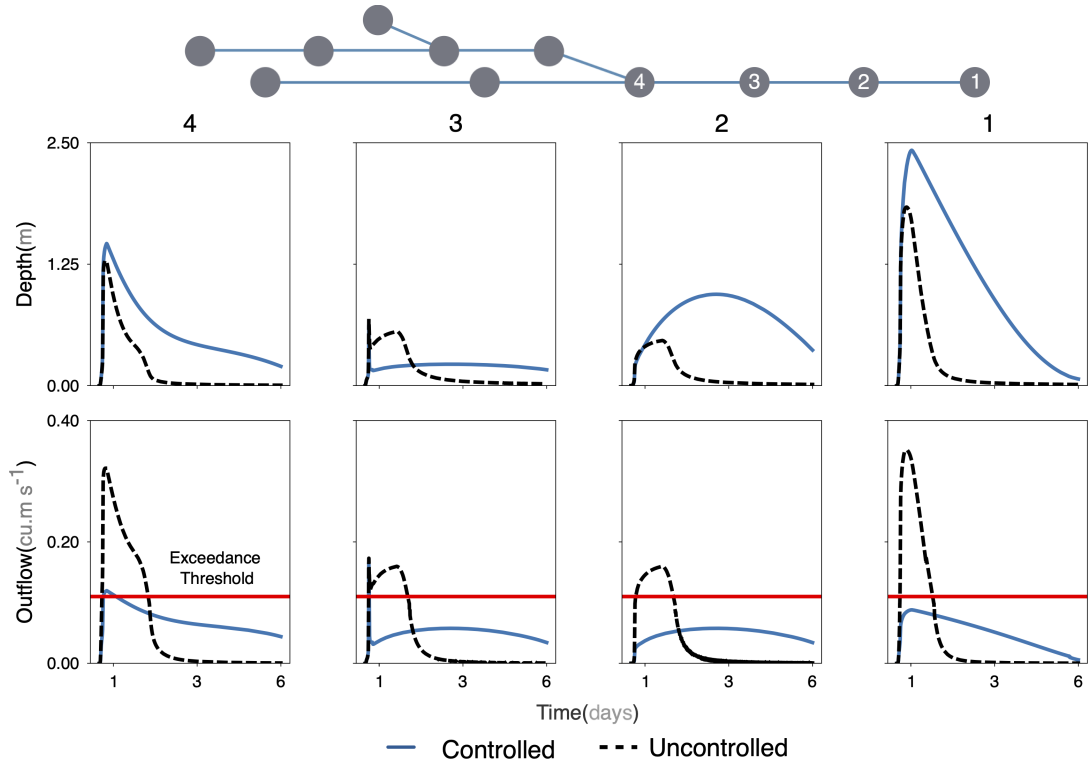}
	\caption{As should be expected, tuning the dimension of a basin's outlet (turning a valve until a desired outlet opening percentage is reached) ahead of a storm event reduces the intensity of its outflows.
	BO identifies, given an incoming storm, the valve positions that can be set in the network to maintain the outflows from the basins below the desired exceedance threshold.
	The network topology illustrates the location of the four controlled basins in the stormwater network.}\label{fig:gamma-4}
\end{figure*}

The performance in scenario gamma, a separated stormwater network, is presented in Fig.~\ref{fig:gamma-4}.
By changing the valve's position at the outlet of the basins to a pre-computed value and maintaining it throughout the storm event, the controller maintains the flows (second row in Fig.~\ref{fig:gamma-4}) in the network below the exceedance threshold.
In this particular scenario, BO identifies 0.34, 1.00, 0.21, 0.20 as the  valve positions for the four basins for maintaining the outflows below $0.11 m^3\sec^{-1}$. These actions reduce the exceeding flows by 99.87\%.
The application of BO control increases the utilization of the storage in the stormwater network (first row in Fig.~\ref{fig:gamma-4}) and reduces the intensity with which runoff leaves the basin.

\

\begin{figure*}
	\includegraphics[width=0.80\linewidth]{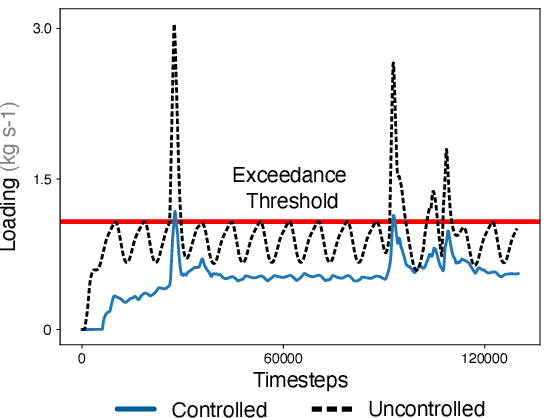}
	\caption{By configuring the inflatable storage dams (ISDs) in a combined sewer network to a pre-defined position, the loading received by the treatment plant, during both dry and wet weather conditions, is maintained below the exceedance threshold. 
	BO, based on the daily flows and incoming storm, identifies the ISD settings that realize this desired response.}\label{fig:epsilon}
\end{figure*}

Fig.~\ref{fig:epsilon} illustrates the controlled and uncontrolled response of scenario epsilon.
BO identifies the percent opening of inflatable storage dams, in the underground pipes of the combined stormwater network, to maintain the loading from the network to the water treatment plant below the desired threshold.
These setpoints\footnote{ \{6\%, 84\%, 92\%, 66\%, 69\%, 73\%, 7\%, 49\%, 1\%, 35\%, 4\% \} } are maintained for a simulated duration of 15 days, during which the network experiences both dry and wet weather flows.
These actions reduce the total loads (above $1.075\ kg\ s^{-1}$) at the treatment plant by 99.11\% (from $7721.27\ kg$ to $68.26\ kg$).
By reducing the outlet pipe dimensions, we increase the time spent by runoff in the pipe network, which directly corresponds to an increased pollutant capture, thus effectively reducing the load leaving the network~\cite{Troutman_2020}.


\begin{table}
\caption{Solutions identified by BO, for the evaluated scenarios, consistently outperform the solutions identified by GA for a set of 30 iterations across 30 random seeds.
The mean and standard deviation of the identified optimal control decision's performance is summarized in the table; lower the values indicate better the ability of the control decision to achieve the desired control objective.
}\label{tab:boga}
\begin{tabular}{p{0.5in}p{0.5in}p{0.5in}p{0.5in}p{0.5in}}
\toprule
\multirow{3}{*}{\textbf{Scenario}} & \multicolumn{2}{c}{\multirow{3}{*}{\textbf{BO}}} & \multicolumn{2}{c}{\multirow{3}{*}{\textbf{GA}}} \\
&    &  &  &   \\
&   $\mu$ & $\sigma$ & $\mu$ &  $\sigma$ \\ \midrule
\texttt{gamma}   & 1875.89 & 961.12 & 3070.28 & 2063.03 \\\midrule
\texttt{epsilon}  & 1008.35 & 573.83 & 2619.65 & 1351.98 \\
\bottomrule
\end{tabular}
\end{table}

\

Table.\ref{tab:boga} summarizes the BO and GA-based controller's performance for scenarios gamma and epsilon for 30 iterations across 30 random seeds.
Performance of the solutions for scenario gamma and scenario epsilon are quantified using Eq.~\ref{eq:gen-gamma-perf} and Eq.~\ref{eq:gen-epslion-perf}, respectively.
BO consistently outperforms the GA in terms of computational efficiency.
BO's average performance for scenarios gamma and epsilon is 63.67\% and 159.79\% better than GA\@. 
Furthermore, BO's performance variance (columns 4 and 6 in Table.\ref{tab:boga}) is considerably lower than GA(by $114.6\%$ and $135.6\%$).

\

In this Generalizability evaluation, the proposed BO approach identifies the control decisions that realize the desired control response, directly ``off-the-shelf'' without any scenario-specific customizations for the stormwater network's topology or the control objective.
This ability to be applied out of the box to control diverse stormwater networks illustrates the Generalizability of the BO approach.
While GA is also generalizable in similar terms, BO has demonstrably been more computationally efficient for the evaluated scenarios. 

\subsection{Uncertainty Quantification}

\begin{figure*}
	\includegraphics[width=\linewidth]{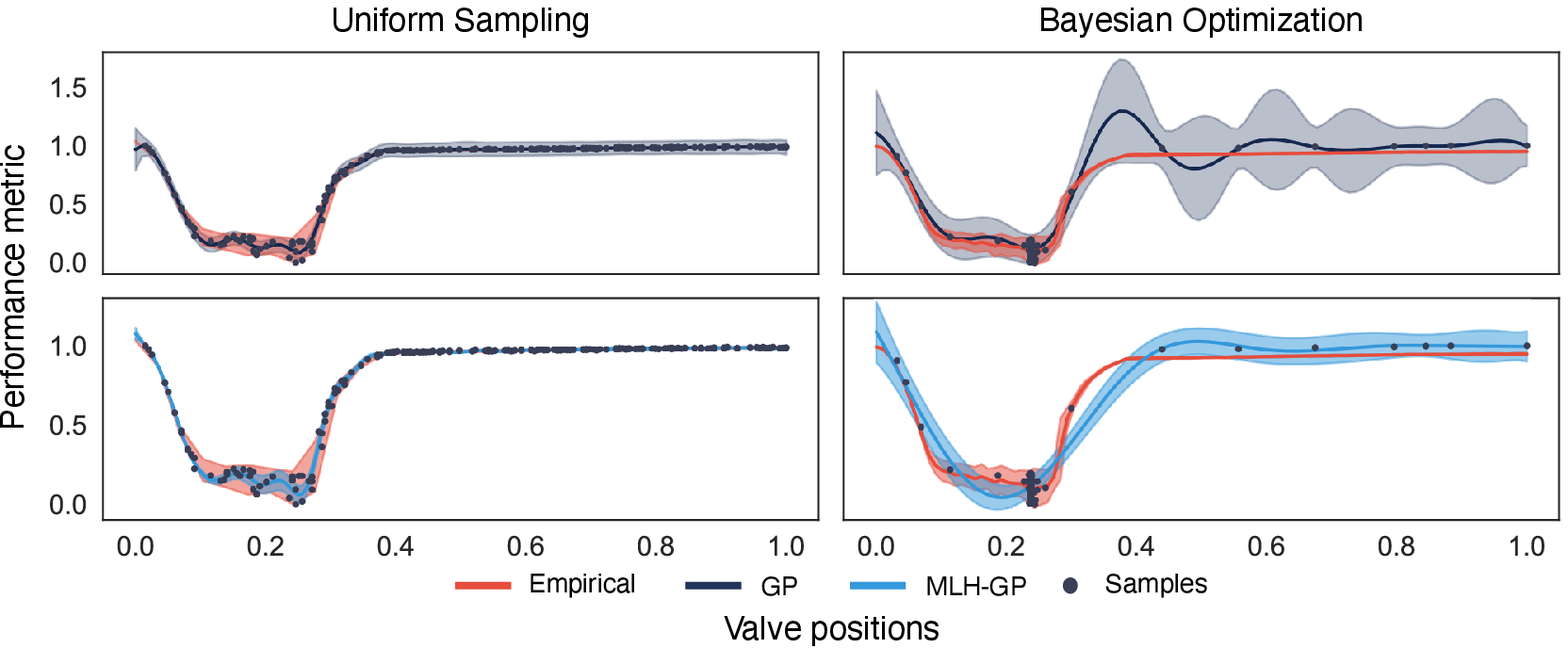}
	\caption{MLH-GP more accurately quantify the manifestation of rainfall uncertainty (shaded areas) in the performance metric than the GP. 
	For the same number of samples (200), uncertainty bounds quantified by MLH-GP closely aligns with the empirically computed values.
	Though GP performs better when the entire solution space is uniformly sampled, they still overestimate the uncertainty compared to MLH-GP.
	Note that most of the 200 samples in BO are focused around 0.23, as acquisition function prioritizes evaluating the most promising regions in the solution space.}\label{fig:uc-limits}
\end{figure*}

Fig.~\ref{fig:uc-limits} illustrates the uncertainty quantified by the BO (blue shaded area) and compares it to the empirically computed uncertainty (orange shaded area) for a single basin in scenario theta.
Though the uncertainties quantified by the BO does not precisely correspond to the empirically estimated values, they do closely align with them.
BO uses ten times fewer samples than used for empirical estimation and assumes a specific kernel\footnote{Squared Expotential Kernel} to represent the relationship between the control decisions and their corresponding uncertainties.
Thus, it is intractable for the estimated uncertainty to exactly converge onto the empirically computed values.
Having said that, MLH-GP much more accurately captures the underlying uncertainty than the GP (gray shaded area), which overestimates the uncertainty  (first row second column in Fig.~\ref{fig:uc-limits}).
While GPs better estimate the uncertainty when the entire solution space is sampled uniformly, they still overestimate (gray area in the first row first column in Fig.~\ref{fig:uc-limits}) the uncertainty when compared to MLH-GP\@.
Uncertainty quantified by MLH-GP almost perfectly aligns with the empirical estimates (~0.3 to 1.0 in second row first column in Fig.~\ref{fig:uc-limits}) for the same set of uniformly sampled data-points.
These results demonstrate the effectiveness of MLH-GP in quantifying the input-dependent uncertainties. 

\

Quantifying the uncertainty associated with control decisions enables us to understand the corresponding risk associated with implementing these decisions in a stormwater network. 
For instance, consider the uncertainty estimates presented in Fig.~\ref{fig:uc-limits}.
For the given set of synthetic rainevents, uncertainty estimates for the valve positions between $\approx0.10$ and $\approx0.25$ (10\% and 25\% valve opening) are slightly larger than those for valve position greater than 0.25.  
From these estimates, we can interpret that a control decision between 0.19 to 0.25 will realize the objective of maintaining the flows below $1.0 m^3 s^{-1}$ and its ability to do, no matter which storm event it experiences, can only deviate by at most 0.076 of the expected performance metric.
Whereas any valve opening greater than 0.25 will not be able to realize the control objective.

\

Understanding the risks associated with control actions enables us to identify the set of actions that can be implemented with a high degree of certainty to achieve the desired control objective, and BO, though the objective function learned by the GP, can be adopted to quantify these risks. 
Furthermore, this objective function also enables us to analytically interpret how the simulated data influences the  choice of optimal actions.

\section{Discussion}
BO identifies control solutions solely based on the specific performance metrics, without the need for internal optimizer parameterization.
Given the sole reliance on one metric, designing a performance metric that accurately and uniquely represents the desired controlled response is paramount.
Building on our previous work~\cite{Mullapudi_Lewis_Gruden_Kerkez_2020}, this paper's performance metrics were designed to accurately represent the desired control objective, to ensure that the BO will not be able to game the performance metric and identify a solution that minimizes the performance metric but results in an undesirable response in the stormwater network.
When adopting BO to control new stormwater systems, performance metrics can be designed using any mathematical structure\footnote{Equations used in the performance metric can be exponential, polynomial, etc\ldots} as long as it accurately represents the desired responseand there does not exist a trivial solution that the BO can exploit.
Performance metrics presented in this work can directly be translated for controlling other stormwater networks for the specific set of control objectives discussed in the paper:maintaining flows and loading below a threshold.

\

Given the complexity and scale of stormwater systems, we cannot guarantee global optimality without enforcing the assumption of linearity. 
Hence, we have to contend with the sub-optimal solutions that achieve the objective. 
For the scenarios evaluated in this paper, BO was successfully able to identify the control decision that achieves the desired response.
However, the quality of the identified solution has a strong dependence on the random seed used.
Hence, readers should consider evaluating BO across multiple random seeds.

\

Results presented in this paper suggest that a sub-set of control objectives, like the ones discussed in this paper, can be realized without real-time control, but rather solely by tuning valve openings ahead of a storm. 
However, it might not often be the case that a high-fidelity model and a representative estimate of incoming storm events would be available. 
In such instances, one could consider a hybrid controller that configures the control elements before the storm event and during the storm event monitors the network's response.
If it detects a deviation from the planned response, it can then implement appropriate control actions in the network.
Such a hybrid control approach would enable us to realize desired control responses with minimal and risk-averse control interventions, and it should be evaluated in future studies.

\

There is an active interest in the AI community in developing Acquisition functions that are robust to the uncertainties inherent in the evaluations~\cite{letham2019}.
These advances stand to improve the BO's ability to quantify the underlying uncertainty in the system. 
Furthermore, these new Acquisition functions improve on the approach's effectiveness by parallelizing the acquisition process~\cite{frazier2018tutorial}. 
Recent advances in Deep-Gaussian processes have demonstrated their effectiveness in working around GP's parametric limitation and would be instrumental in extending this approach for the control of larger systems~\cite{damianou2013deep}.
Though promising, their applicability for the control of stormwater systems is yet to be evaluated. 

\

Understanding the impacts of uncertainties inherent in stormwater systems would be essential in developing robust stormwater control algorithms.
BO, as illustrated in this paper, is an efficient approach for quantifying these uncertainties. 
Though the GP can quantify the uncertainties based on the observed data points\footnote{Though in this paper, they are simulated, this approach can be adopted for real-world data. Hence, to illustrate the flexibility of the approach, we refer to them as observed data points.}, interpreting and translating them into actionable information for stormwater control decision-making is still an open research question.
Especially in the instances where multiple stormwater basins are being controlled (e.g.\ 10, 20 basins), interpreting these high-dimensional uncertainty estimates is a non-trivial task.
To the best of our knowledge, there does yet exist an approach for tackling this, and addressing this knowledge gap would be instrumental in transitioning stormwater control algorithms into adoption. 

\section{Conclusion}
This paper introduces a BO-based automated control algorithm for identifying control decisions that realize the desired control objective.
To the best of our knowledge, this is the first application of BO for the control of stormwater systems.
This algorithm also incorporates a methodology for quantifying the rainfall uncertainties associated with a control decision's performance.
Though this control approach can be used off-the-shelf for controlling stormwater systems, it is limited in the set of control objectives that it can achieve.
Its ability to identify the control decisions that realize the desired control objective depends on the performance metric representing the objective and the random seed used in the methodology.
The source code accompanying this work should allow researchers to evaluate the BO-based control approach's performance on other stormwater systems and develop extensions for improving its effectiveness.

\bibliographystyle{ACM-Reference-Format}
\bibliography{reference}

\end{document}